\documentclass[letterpaper, 10 pt, conference]{ieeeconf}  

\IEEEoverridecommandlockouts                              

\usepackage{cite}
\usepackage{booktabs}
\usepackage{amsmath,amssymb,amsfonts}
\usepackage{graphicx}
\usepackage{textcomp}
\usepackage{duckuments}
\usepackage{xcolor}
\usepackage{soul}
\usepackage{subcaption}
\usepackage[ruled,linesnumbered, vlined]{algorithm2e}
\DontPrintSemicolon
\SetCommentSty{myCommentStyle}

\usepackage{amsfonts}
\let\labelindent\relax
\usepackage{enumitem}
\usepackage{amsmath}
\usepackage{graphicx}
\usepackage{subfig}
\usepackage[margin=0.792in]{geometry}
\usepackage{bbm}
\usepackage{algpseudocode} 
\usepackage{multirow}
\usepackage{xcolor}
\usepackage{ragged2e}

\def\BibTeX{{\rm B\kern-.05em{\sc i\kern-.025em b}\kern-.08em
    T\kern-.1667em\lower.7ex\hbox{E}\kern-.125emX}}

\title{\LARGE \bf Physics-informed Neural Motion Planning via Domain Decomposition in Large Environments}

\author{Yuchen Liu, Alexiy Buynitsky, Ruiqi Ni, and Ahmed H. Qureshi
\thanks{Yuchen Liu, Alexiy Buynitsky, Ruiqi Ni, and Ahmed H. Qureshi are with the Department of Computer Science, Purdue University, West Lafayette, IN, USA, 47907. Email {\tt\small$\{$liu3853, abuynits, ni117, ahqureshi$\}@$purdue.edu}}%
}

\begin{document}


\let\oldtwocolumn\twocolumn
\renewcommand\twocolumn[1][]{%
    \oldtwocolumn[{#1}{
    \begin{center}
    \vspace{-5mm}
    \includegraphics[width=\textwidth]{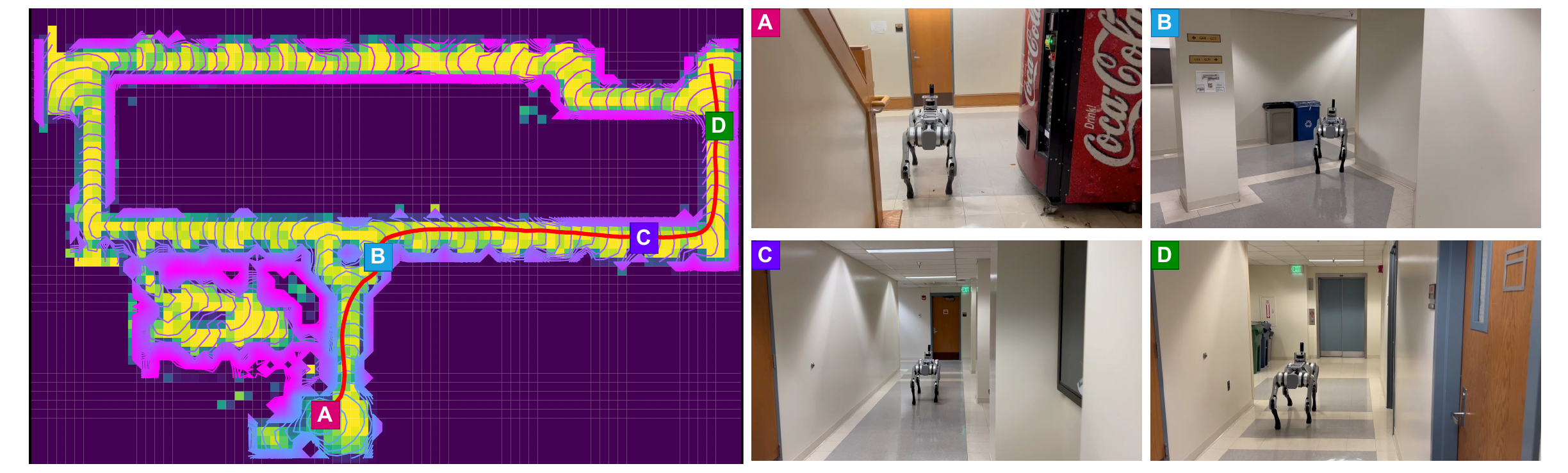}
    \captionof{figure}{\justifying Demonstration of our approach in real-world indoor HAAS scenario with 843 Sq. meters. The robot walks through multiple narrow passages and hallways to reach the destination in front of an elevator. In this particular scenario, our method took 0.15 seconds, while RRTConnect and LazyPRM took 3.39 and 2.57 seconds, respectively, to find their path solution.}
           \label{fig:haas}
        \end{center}
    }]
}

\maketitle
\thispagestyle{empty}
\pagestyle{empty}

\begin{abstract}
Physics-informed Neural Motion Planners (PiNMPs) provide a data-efficient framework for solving the Eikonal Partial Differential Equation (PDE) and representing the cost-to-go function for motion planning. However, their scalability remains limited by spectral bias and the complex loss landscape of PDE-driven training. Domain decomposition mitigates these issues by dividing the environment into smaller subdomains, but existing methods enforce continuity only at individual spatial points. While effective for function approximation, these methods fail to capture the spatial connectivity required for motion planning, where the cost-to-go function depends on both the start and goal coordinates rather than a single query point. We propose Finite Basis Neural Time Fields (FB-NTFields), a novel neural field representation for scalable cost-to-go estimation. Instead of enforcing continuity in output space, FB-NTFields construct a latent space representation, computing the cost-to-go as a distance between the latent embeddings of start and goal coordinates. This enables global spatial coherence while integrating domain decomposition, ensuring efficient large-scale motion planning. We validate FB-NTFields in complex synthetic and real-world scenarios, demonstrating substantial improvements over existing PiNMPs. Finally, we deploy our method on a Unitree B1 quadruped robot, successfully navigating indoor environments. The supplementary videos can be found at https://youtu.be/OpRuCbLNOwM.
\end{abstract}
\section{Introduction}
\label{sec:introduction}
Motion planning is a fundamental problem in robotics \cite{lavalle2006planning}, aiming to compute a collision-free path between a given start and goal while satisfying constraints. Classical approaches, such as sampling-based methods \cite{karaman2011sampling,lavalle2001rapidly,gammell2014informed,janson2015fast}, offer theoretical guarantees but can be computationally expensive in high-dimensional spaces. Learning-based methods \cite{qureshi2019motion,qureshi2020motion,ichter2018learning,qureshi2018deeply,kumar2019lego,chaplot2021differentiable} improve efficiency by leveraging prior experience but rely heavily on expert demonstrations or reinforcement learning, limiting their adaptability to new environments.

Recent advances in PiNMPs \cite{ni2023ntfields,ni2023progressive,shen2024pc}, such as Neural Time Fields (NTFields) \cite{ni2023ntfields} and Progressive NTFields (P-NTFields) \cite{ni2023progressive}, have introduced a data-efficient alternative by solving the Eikonal Partial Differential Equation (PDE) directly using Physics-Informed Neural Networks (PINNs) \cite{raissi2019physics,karniadakis2021physics}. These methods learn implicit cost-to-go functions without requiring expert data. NTFields solve the non-linear Eikonal PDE, while P-NTFields introduce a viscosity term \cite{crandall1983viscosity} to stabilize learning through a reformulated semi-linear elliptic PDE. Despite their efficiency, these approaches struggle with scalability due to two key limitations: spectral bias and the complex loss landscape of PDE-driven training.

Spectral bias refers to the tendency of neural networks to learn low-frequency features more easily than high-frequency ones \cite{xu2019frequency}. In large environments, due to input normalization, previously low-frequency features shift into a high-frequency regime, making them harder to learn. Meanwhile, PDE-driven training induces highly complex optimization landscapes \cite{krishnapriyan2021characterizing}, further complicating convergence as environment size increases. These challenges limit the practical scalability of PiNMPs, making them infeasible for large-scale motion planning.

To address these challenges, domain decomposition has emerged as a promising approach \cite{moseley2023finite}. Inspired by Finite Element Methods (FEMs), it divides the problem into smaller subdomains, solving them independently before assembling a global solution. While this mitigates spectral bias and improves convergence, existing methods—such as KiloNeRF \cite{reiser2021kilonerf}—focus on mapping individual spatial coordinates to scalar fields while enforcing continuity only at single points. Although effective for static function approximation, these methods fail to capture broader spatial connectivity, which is critical for motion planning, where the cost-to-go function depends on both the start and goal coordinates, not just a single spatial query.

To overcome this limitation, we introduce FB-NTFields, a novel neural field representation designed for scalable motion planning. Instead of enforcing continuity in the output space, FB-NTFields construct a latent space representation, mapping spatial locations into a structured latent space. The cost-to-go function is then computed as a distance between the latent embeddings of the start and goal coordinates, ensuring spatial coherence across the environment. Additionally, FB-NTFields integrate domain decomposition, partitioning the environment into overlapping subdomains to capture localized frequency components while maintaining global consistency in the latent space.

We evaluate FB-NTFields across complex synthetic and real-world environments, demonstrating substantial improvements over NTFields and P-NTFields, which struggle to scale effectively. Furthermore, we benchmark our method against traditional motion planning approaches, showing that FB-NTFields achieve lower planning times while maintaining high success rates. Finally, we deploy FB-NTFields on a Unitree B1 quadruped robot, successfully navigating complex indoor environments, shown in Fig. \ref{fig:haas}.

\section{Related Work}
\label{sec:related}

The development of efficient and scalable motion planning methods has been an active research area for decades. Among traditional approaches, the Fast Marching Method (FMM) \cite{sethian1996fast} is a notable technique for solving the Eikonal PDE via discretization and search. However, FMM lacks a continuous representation of the solution and becomes computationally intractable in high-dimensional spaces. Other classical approaches include sampling-based motion planning \cite{karaman2011sampling,lavalle2001rapidly,gammell2014informed,janson2015fast}, which offer probabilistic completeness but suffer from high computational costs, and trajectory optimization \cite{von1992direct,betts1993path,zucker2013chomp}, which produces smooth paths but is prone to local minima.

To improve efficiency, learning-based motion planning methods have gained attention. Imitation learning-based approaches \cite{qureshi2019motion,qureshi2020motion,ichter2018learning,qureshi2018deeply,kumar2019lego,chaplot2021differentiable} train models to infer paths quickly in complex environments using expert demonstrations. However, they require large-scale expert data, which is computationally expensive to generate. Reinforcement learning (RL)-based methods \cite{tamar2016value,faust2018prm,srinivas2018universal,lawson2023control} enable adaptive decision-making but suffer from high sample complexity and difficulty enforcing collision constraints, limiting their applicability in motion planning.

A recent class of methods, PiNMPs \cite{ni2023ntfields,ni2023progressive,ni2024physics,shen2024pc}, leverages PINNs \cite{raissi2019physics,karniadakis2021physics} to solve the Eikonal PDE directly. Unlike learning-based approaches that require expert data, PiNMPs enforce a PDE-driven loss function based on the gradient consistency condition of the Eikonal equation. The motion behavior is determined by a speed field, where higher speeds correspond to free space and lower speeds indicate proximity to obstacles. This formulation allows PiNMPs to infer cost-to-go representations efficiently using only randomly sampled robot configurations and distance-to-obstacle information.

Once trained, PiNMPs provide rapid path inference, as their learned arrival time field enables gradient-based backtracking to extract paths. These methods have demonstrated superior efficiency over imitation learning-based planners, offering training without demonstration trajectories and real-time inference. However, their scalability remains limited due to PDE-driven optimization complexity and spectral bias—a phenomenon where neural networks struggle to learn high-frequency details in large environments \cite{moseley2023finite}. As the environment size increases, low-frequency features shift into high-frequency regimes, further exacerbating training difficulties.

To improve the scalability of PINNs and neural fields, domain decomposition has been widely explored \cite{moseley2023finite, reiser2021kilonerf, chen2022tensorf, muller2022instant}. Inspired by FEMs, domain decomposition divides the problem into smaller subdomains, solving them independently before assembling a global solution. This technique reduces spectral bias and optimization complexity, making it a promising strategy for large-scale problems.

Neural field-based methods, such as NeRF-inspired models \cite{mildenhall2021nerf}, have adopted similar decomposition techniques, leveraging small local MLPs \cite{reiser2021kilonerf}, tensor decomposition \cite{chen2022tensorf}, and hash-based acceleration \cite{muller2022instant} to improve efficiency. While these methods enable faster inference, they enforce continuity only at discrete spatial points, making them unsuitable for motion planning, where solutions must depend on both start and goal coordinates rather than isolated queries. Furthermore, existing decomposition methods fail to preserve global spatial connectivity, a crucial property for cost-to-go estimation in large environments.

We introduce FB-NTFields, a novel scalable neural field representation that extends domain decomposition to motion planning. By constructing a latent space representation, FB-NTFields compute cost-to-go as a distance between latent embeddings of the start and goal, ensuring spatial coherence while maintaining computational efficiency. Additionally, by integrating overlapping subdomains, FB-NTFields capture local frequency components while preserving global connectivity, overcoming spectral bias and PDE optimization challenges, making them well-suited for large-scale motion planning.
\section{Background}
\label{sec:setup}
This section provides an overview of robot motion planning problems and their solutions via PiNMPs.

Let the robot workspace be denoted by $\mathcal{X} \subset \mathbb{R}^m$, with dimensionality $m \in \mathbb{N}$. The workspace includes the obstacle and obstacle-free space denoted by $\mathcal{X}_{obs} \subset \mathcal{X}$, and $\mathcal{X}_{free} = \mathcal{X} \setminus \mathcal{X}_{obs}$, respectively. Let the robot's configuration space (C-space) be denoted as $\mathcal{Q} \subset \mathbb{R}^d$, where $d \in \mathbb{N}$ represents degree-of-freedom. Within this space, the obstacle and obstacle-free space are represented by $\mathcal{Q}_{obs} \subset \mathcal{Q}$ and $\mathcal{Q}_{free} = \mathcal{Q} \setminus \mathcal{Q}_{obs}$, respectively. 

The motion planning problem is to find an obstacle-free trajectory $\boldsymbol{\tau} \subset \mathcal{Q}_{free}$ comparing configuration sequence connecting the given start $\boldsymbol{q}_s$ and goal $\boldsymbol{q}_g$. There exist a variety of ways to solve motion planning problems. One of those ways is to solve the Eikonal PDE that relates the speed $S(\boldsymbol{q}_g)$ at goal configuration, $\boldsymbol{q}_g$, to arrival time field, $T(\boldsymbol{q}_s,\boldsymbol{q}_g)$, between the given start, $\boldsymbol{q}_s$, and goal, $\boldsymbol{q}_g$, i.e.,
\begin{equation}
\frac{1}{S(\boldsymbol{q}_g)} = \|\nabla_{\boldsymbol{q}_g} T(\boldsymbol{q}_s, \boldsymbol{q}_g)\|,
\label{eikonal}
\end{equation}

The solution of the Eikonal equation can be expressed as the travel time $T(\boldsymbol{q}_s, \boldsymbol{q}_g)$,  which represents the time required for a wavefront with speed $S(\boldsymbol{q}_g)$ to propagate from the start $\boldsymbol{q}_s$ to the goal $\boldsymbol{q}_g$. The shortest path can then be obtained by following the negative gradient of the travel time field. Recent advancements have introduced PINNs, such as NTFields and P-NTFields \cite{ni2023ntfields,ni2023progressive}, which solve the Eikonal PDE for robot motion planning. However, we highlight that NTFields and P-NTFields struggle with scalability in large environments, where the prevalence of high-frequency features exacerbates the challenges posed by spectral bias, leading to performance degradation.


\begin{figure*}[t]
\centering
\vspace{2.5mm}
    \includegraphics[width=1.0\textwidth]{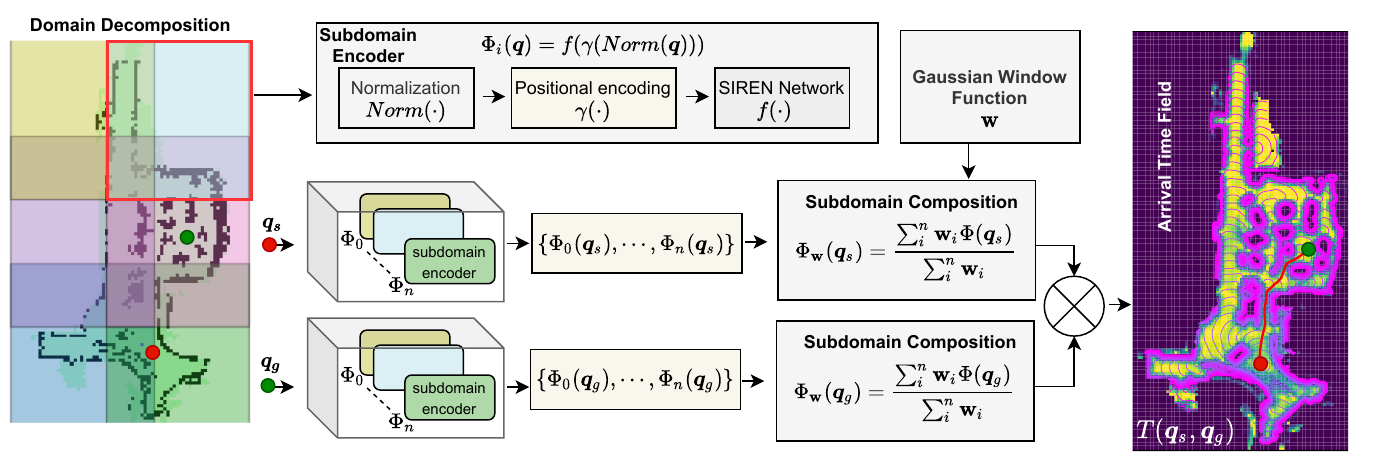}
    \caption{FB-NTFields workflow decomposes the given domain into overlapping subdomains. The given start, $\boldsymbol{q}_s$, and goal, $\boldsymbol{q}_g$, configurations are encoded by the subdomain encoders $\{\Phi_0,\cdots, \Phi_n\}$. These embeddings are then composed together into latent embeddings, $\Phi_\mathbf{w}(\boldsymbol{q}_s)$ and $\Phi_\mathbf{w}(\boldsymbol{q}_g)$, using the weighting function $\mathbf{w}$. These global embeddings are passed through a time field generator, $\bigotimes$, to predict the arrival time, $T(\boldsymbol{q}_s, \boldsymbol{q}_g)$, between the given start and goal. The gradients of the predicted arrival time field computed via backpropagation with respect to start and goal provide their respective speed fields. The maps illustrated in this figure belong to a real-world scenario with 1859.6 Sq. meters of dimension.}
    \label{fig:pipeline}
\vspace{-4.5mm}
\end{figure*}

\section{Method: Finite Basis NTFields}
\label{sec:methods}
 This section introduces FB-NTFields, illustrated in Fig. \ref{fig:pipeline}. Unlike conventional methods that directly learn the output signal space, FB-NTFields operate in a latent space representation, enhancing scalability and generalization for large-scale motion planning. Following a divide-and-conquer strategy \cite{moseley2023finite, reiser2021kilonerf, chen2022tensorf, muller2022instant}, our approach partitions the domain into multiple overlapping subdomains, each assigned a dedicated neural network that learns a local latent representation instead of directly predicting the solution. These latent features are then seamlessly integrated into a globally coherent representation, enabling efficient and scalable optimization of the Eikonal PDE for motion planning. While our method is designed for large environments and applicable to mobile robots, we focus on workspace coordinates in the configuration space $\mathcal{Q}$ for simplicity. 
The key components of our approach are detailed below.
\subsection{Domain Decomposition}
We divide a domain $\mathrm{Q}$ into $n\in \mathbb{N}$ overlapping subdomains $\Omega=\{\boldsymbol{\omega}_0,\cdots,\boldsymbol{\omega}_n\}$, with their associated boundaries $\{\boldsymbol{b}_0,\cdots, \boldsymbol{b}_n\}$. The division could be in any regular or irregular shape with overlapping neighboring parts. This overlapping is critical to ensure differentiability at the subdomain boundaries and globally optimize the PDE loss. An example division of a domain $\mathrm{Q}$ is shown in Fig. \ref{fig:pipeline}. We also discard subdomains that fall entirely inside obstacles since modeling the arrival time field inside obstacle space is not required for motion planning. Discarding these subdomains also contributes to reducing the number of subdomain encoders and overall neural parameters.


\subsection{Subdomain Encoder}
The subdomain encoder takes a configuration sample $\boldsymbol{q}$ and provides its subdomain-specific latent embedding $\Phi (\boldsymbol{q})$. Given a configuration sample and a subdomain $\omega_i$ with boundary $\boldsymbol{b}_i$, we normalize the configuration sample according to the domain boundary as: 
\begin{equation}
\begin{aligned}
    \boldsymbol{\hat{q}}= Norm(\boldsymbol{q}) = \frac{\boldsymbol{q} - \boldsymbol{b}^{min}_i}{\boldsymbol{b}^{max}_i-\boldsymbol{b}^{min}_i} - {0.5}
    \end{aligned}
    \label{normalize}
\end{equation}
Next, we obtain the Fourier features representing positional encoding of the normalized configuration \cite{tancik2020fourier}, i.e.,
\begin{equation}
\begin{aligned}
    &\gamma(\boldsymbol{\hat{q}})=[\cos(2\pi \mathbf{c}^\top \boldsymbol{\hat{q}}),\sin(2\pi \mathbf{c}^\top \boldsymbol{\hat{q}})]
    \end{aligned}
    \label{pe}
\end{equation}
where $\mathbf{c} \in \mathbb{R}^p$ is a $p \in \mathbb{N}$ dimensional, fixed random latent code. These positional encodings are then passed through a SIREN neural network $f$ \cite{sitzmann2019siren}, which is a multi-layer perceptron with the Sine activation function. The SIREN network has been demonstrated to capture the high-frequency features better and allow smooth differentiation for gradient computation and PDE loss optimization. In summary, we obtain a subdomain latent embedding of a given configuration $\boldsymbol{q}$ as $\Phi(\boldsymbol{q}) = f(\gamma(Norm(\boldsymbol{q})))$. The center of each subdomain and its encoder is aligned so that the subdomain center is fully represented by its own network, whilst the overlapping region is represented by the sum of the overlapping networks.
\subsection{Global Encoding: Subdomain Composition}
For each configuration, we obtain subdomain encodings across all subdomains. This leads to $n$ encodings of a given configuration, i.e., $\Phi_\Omega(\boldsymbol{q})=\{\Phi_0(\boldsymbol{q}), \cdots, \Phi_n(\boldsymbol{q})\}$. These embeddings are composed together using a weighted {average}:\vspace{-0.00in} 
\begin{equation}
\begin{aligned}    
\Phi_{\mathbf{w}}(\boldsymbol{q}) = \frac{\sum^n_i \mathbf{w}_i \Phi(\boldsymbol{q})}{\sum^n_i \mathbf{w}_i},
\end{aligned}
\label{unnormalize}
\end{equation}
where {$\mathbf{w}_i$} is weightage of subdomain $i \in [0,n]$. The weight function can be any function as long as its value is non-zero in the sample's associated subdomains and zero outside those subdomains. In our case, we choose a Gaussian weight function with mean being the center of the subdomain and variance spanning until the subdomain boundary. Therefore, if a sample, $\boldsymbol{q}$, is near the center of a subdomain, it will have only one embedding from that subdomain, while embeddings from other subdomains will have zero weightage. Similarly, when a sample is near the subdomain boundary, it would have a contribution from neighboring subdomains whose embeddings will be included based on their Gaussian weight. Hence, samples in the overlapping regions have weighted encodings from all contributing subdomains.    
\subsection{Composite Time Field Generator}
In this section, we present our procedure to obtain the arrival time field $T(\boldsymbol{q}_s,\boldsymbol{q}_g)$ between a given start, $\boldsymbol{q}_s$, and goal, $\boldsymbol{q}_g$, configurations. For these configurations, we begin by obtaining their weighted subdomain encodings $\Phi_\mathbf{w}(\boldsymbol{q}_s) \in \mathbb{R}^k$ and $\Phi_\mathbf{w}(\boldsymbol{q}_g) \in \mathbb{R}^k$, where $k$ is the embedding dimension. Next, we compute the LogSumExp of the absolute value of the element-wise difference between those two latent encodings. This indicates the smooth $L_\infty$ norm in the latent space to compute the arrival time $T$, i.e.,
\begin{equation}
\begin{aligned}
T(\boldsymbol{q}_s, \boldsymbol{q}_g) &=  \Phi_{\mathbf{w}}(\boldsymbol{q}_s) \bigotimes \Phi_{\mathbf{w}}(\boldsymbol{q}_g) \\&=  \alpha \cdot \log\left( \sum \exp\left( \beta \cdot |\Phi_{\mathbf{w}}(\boldsymbol{q}_s) - \Phi_{\mathbf{w}}(\boldsymbol{q}_g)| \right) \right)
\end{aligned}
\label{symop}
\end{equation}

The hyperparameters $\alpha$ and $\beta$ serve as scaling factors. Notably, our formulation preserves the symmetry of the arrival time field, meaning that the arrival time remains identical whether measured from the start to the goal or vice versa. Consequently, in Eq.~\ref{symop}, swapping $\boldsymbol{q}_s$ and $\boldsymbol{q}_g$ still produces the same prediction for $T$, thereby maintaining the symmetric property within our network.

Our time field generator also draws inspiration from the symmetric operator introduced in NTFields \cite{ni2023ntfields} to enforce this property. In NTFields, the start and goal configurations are encoded by a shared MLP, and these encodings are symmetrized using operators such as the maximum and minimum of their differences. In our architecture, we adopt a similar strategy by employing the $L_\infty$ norm of the difference between encodings, with the distinction that our configuration samples are processed by their respective subdomain encoders.

\subsection{Training Objectives}
We train our subdomain encoders and time field generator end-to-end using the following objective function.
\begin{equation}
\begin{aligned}
L(S^*,S)=    &(\sqrt{S(\boldsymbol{q}_s)/S^*(\boldsymbol{q}_s)}-1)^2+\\
    &(\sqrt{S(\boldsymbol{q}_g)/S^*(\boldsymbol{q}_g)}-1)^2
\end{aligned}
\label{trainloss}
\end{equation}
The $S$ in the above loss function is the predicted speed. It is computed using the Eikonal Equation (Eq.~\ref{eikonal}). Our time field generator predicts the arrival time $T(\boldsymbol{q}_s, \boldsymbol{q}_g)$ between the given start and goal. The predicted $T(\boldsymbol{q}_s, \boldsymbol{q}_g)$ and its gradient with respect to $\boldsymbol{q}_s$ and $\boldsymbol{q}_g$ parameterize the Eikonal Equation, which leads to the predicted speeds $S(\boldsymbol{q}_s)$ and $S(\boldsymbol{q}_g)$, respectively. The ground truth speed function is composed of a truncated signed distance function ($TSDF$), $S^*(\boldsymbol{q})=(TSDF(\boldsymbol{q}))^2(2-TSDF(\boldsymbol{q}))^2$. The truncated signed distance function is as follows:
\begin{equation}
TSDF(\boldsymbol{q}) = \frac{s_{const}}{d^{max}_{obs}} \times \mathrm{clip}(\boldsymbol{\mathrm{d}}_{obs}(\boldsymbol{q}, \mathcal{X}_{obs}), d^{min}_{obs}, d^{max}_{obs})
\label{speed}
\end{equation}
The $d_{obs}$ provides the distance of a given configuration to the obstacles. The $d^{min}_{obs}$ and $d^{max}_{obs}$ are the minimum and maximum distance thresholds. The $s_{const}$ is a fixed scaling factor. Note that when distance is clipped to max distance, the speed $S^*$ becomes a constant value. This adds an upper bound in optimization PDE loss. The $d^{min}_{obs}$ controls the safety margin of the robot to the obstacles. If $d^{min}_{obs}$ is set to a very small number, then the resulting path solutions will have less distance threshold from the obstacles. Finally, we should also highlight that the only data needed to train our model is the randomly sampled robot start and goal configurations and their distance to obstacles. This type of data is much easier to obtain than the data needed by the existing imitation learning-based method, which requires not only randomly sampled start and goal pairs but also the collision-free trajectories connecting them. 

\subsection{Path Inference}

Once the time field generator is trained, we can directly compute path sequences connecting the given start 
$\boldsymbol{q}_s$ and goal 
$\boldsymbol{q}_g$ by following the negative gradient of the arrival time field. This procedure, similar to NTFields \cite{ni2023ntfields}, enables fast and efficient path extraction. In summary, the path solution is obtained as follows:

\begin{equation}
    \begin{aligned}
    \boldsymbol{q}_s &\gets \boldsymbol{q}_s-\gamma S^2(\boldsymbol{q}_s)\nabla_{\boldsymbol{q}_s} T(\boldsymbol{q}_s,\boldsymbol{q}_g) \\
    \boldsymbol{q}_g &\gets \boldsymbol{q}_g-\gamma S^2(\boldsymbol{q}_g)\nabla_{\boldsymbol{q}_g} T(\boldsymbol{q}_s,\boldsymbol{q}_g)
    \end{aligned}
    \label{plan}
\end{equation}


where $\gamma$ is a step size hyperparameter, and the gradient of the time field provides the direction of motion. The scaling factor $S^2$ dynamically adjusts step sizes—reducing movement near obstacles and increasing step size in free space, ensuring safe and smooth navigation. This behavior is also evident in real-world experiments, where the robot’s motion naturally slows near obstacles, reducing drift and improving trajectory following for the underlying motion controller.


The path-finding process is conducted bidirectionally, iterating for a fixed number of steps or until the Euclidean distance between the updated $\boldsymbol{q}_s$ and $\boldsymbol{q}_g$ falls below a threshold. Since the time field is predicted using Eq.~\ref{symop}, updated configurations may shift into new subdomains. Our subdomain composition encoder dynamically reweights relevant subdomains (Eq.~\ref{unnormalize}), ensuring that the time field generator always receives updated subdomain-weighted embeddings for accurate inference.

Our fast path inference provides a high-quality initial guess significantly faster than sampling-based methods, enabling rapid replanning in dynamic environments. Additionally, the global initial guess can be leveraged by trajectory optimization frameworks to refine paths in the presence of moving obstacles and dynamic constraints.

\subsection{Implementation details} We have implemented our method in PyTorch and plan to open-source our code on GitHub along with the final version of the paper. This will include the configuration files containing hyperparameters and neural network training details for all the experiments presented in this paper.

\section{Evaluation and Discussion}
In this section, we present the experimental evaluation of our proposed method. We begin by comparing our approach with previous methods, highlighting that merely increasing the model size is insufficient for achieving high performance. Next, we benchmark our method against state-of-the-art planning algorithms across ten complex environments from the Gibson dataset \cite{li2021igibson}. Finally, we demonstrate the scalability of our approach in two real-world scenarios: long passage halls and cluttered study areas. All experiments were performed on a system equipped with an RTX 4090 GPU, an Intel Xeon W5-3435X CPU, and 64GB of RAM.

\subsection{Ablation Studies}

\begin{figure}[t]
    \centering
    \begin{minipage}{\columnwidth} 
        \centering

        \includegraphics[width=\columnwidth]{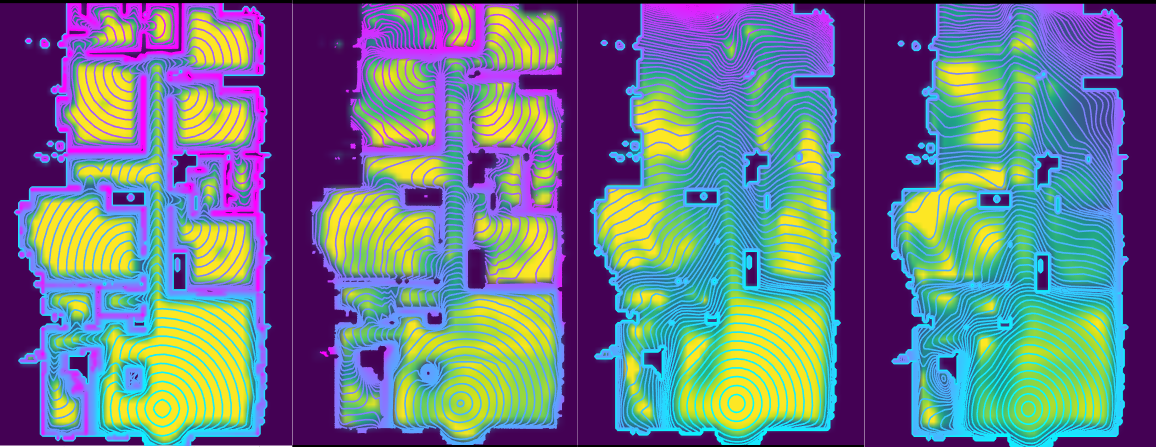}
        \put(-225,95){{ FMM}}
        \put(-161,95){{ Ours}}
        \put(-110,95){{ NTFields}}
        \put(-55,95){{ P-NTFields}}
    \end{minipage}
    \vspace{2mm} 
    
    \begin{minipage}{\columnwidth} 
        \centering
        \resizebox{\columnwidth}{!}{%
        \begin{tabular}{l|c|c|c|c|c|c}
            \hline
            \multirow{2}{*}{Method} & \multicolumn{2}{c|}{Simple (Aloha)} & \multicolumn{2}{c|}{Medium (Hercules)} & \multicolumn{2}{c}{Hard (Calavo)} \\
            \cline{2-7}
            & SR (\%) & ET & SR (\%) & ET & SR (\%) & ET \\
            \hline
            ours\_18k & 98.0 & 0.151 & 95.5 & 0.151 & 89.0 & 0.155 \\
            ours\_41k & 98.5 & 0.304 & 98.5 & 0.294 & 99.0 & 0.264 \\
            ours\_68k & 99.0 & 0.435 & 99.0 & 0.471 & 97.5 & 0.319 \\
            \hline
            nt\_69k & 58.5 & 0.252 & 38.5 & 0.256 & 25.5 & 0.254 \\
            nt\_270k & 73.0 & 0.277 & 40.5 & 0.288 & 32.0 & 0.285 \\
            nt\_441k & 69.0 & 0.311 & 42.5 & 0.312 & 37.5 & 0.310 \\
            \hline
            pnt\_74k & 71.0 & 0.389 & 46.0 & 0.340 & 35.5 & 0.376 \\
            pnt\_282k & 68.0 & 0.780 & 48.0 & 0.820 & 31.5 & 1.041 \\
            pnt\_453k & 66.0 & 1.245 & 45.5 & 1.239 & 30.5 & 1.443 \\ 
            \hline
        \end{tabular}%
        }
    \end{minipage}
    \caption{Ablation comparison across three Gibson environments. The figure displays travel time contour lines for the ground truth, our method, NTFields, and P-NTFields in the hard environment (Calavo). The accompanying table reports the path planning success rate and per-epoch training time in seconds as model sizes increase for all three methods. The success rate is evaluated using 200 randomly selected start-goal configuration pairs in free space.}
    \label{fig:ablation}
    \vspace{-0.1in}
\end{figure}

The primary objective of our ablation studies is to demonstrate that the positional encoding and SIREN networks within the single-MLP architecture employed by NTFields and P-NTFields are inadequate for capturing high-frequency signals. We evaluate our method alongside NTFields and P-NTFields across three Gibson environments—Aloha, Hercules, and Calavo—which correspond to increasing levels of difficulty (Simple, Medium, and Hard). For a fair comparison, model sizes are increased for all methods as illustrated in the table in Fig.~\ref{fig:ablation}. While NTFields and P-NTFields scale by increasing the depth and width of their networks, our approach scales through domain decomposition by increasing the number of subdomains.

The metrics considered in these studies include the success rate (SR) and the epoch time (ET) during training. NTFields and P-NTFields are trained for a fixed 10,000 epochs, while our method converges more rapidly, requiring only 2,000 epochs. Additionally, we also report the number of parameters used by each model. For our method, we report the total number of parameters for the full domain; parameters corresponding to subdomains with no free space are discarded, so the actual count may be lower. Notably, our largest model has fewer parameters than the smallest models of the other two methods. This efficiency is due to two factors: (1) each subnetwork is much smaller, and (2) our design eliminates the need for a decoder network required by NTFields and P-NTFields.  

Experimental results show that our small model (14k parameters) performs well in simple environments but loses accuracy as the complexity increases. Both the medium (41k parameters) and large (68k parameters) models maintain a high success rate across all environments. In contrast, NTFields exhibit only a limited improvement in success rate even with dramatic increases in model size, whereas P-NTFields face difficulties in converging to the right results as the model size increases.

In Fig.~\ref{fig:ablation}, all methods with the highest success rates among all models are compared against the ground truth in the hard environment (Calavo). The background color represents the predicted speed fields, while the contour lines indicate the predicted travel times. Since the negative gradient of the travel times corresponds to the predicted trajectories, it is evident that our method successfully identifies obstacle spaces and finds collision-free paths. In contrast, NTFields and P-NTFields exhibit poor learning outcomes, resulting in trajectories that lead to collisions. Overall, our experiments demonstrate the effectiveness of our approach in complex environments, whereas NTFields and P-NTFields fail to achieve comparable performance even with increased model sizes.

\begin{table}[h]
\centering
\scalebox{1}{ 
\begin{tabular}{cccc}
\toprule
\#& \multicolumn{1}{c}{Env Name} & \multicolumn{1}{c}{Dim Sq. m} & \multicolumn{1}{c}{\# rooms} \\ 
\midrule
1  & Aloha & $111.74$ & 8 \\
2  & Cabin & $155.70$ & 9 \\
3 & Foyil & $90.20 $ & 7 \\
4 & Dalcour & $177.41$ & 11 \\
5 & Hercules & $119.16 $ & 8 \\
6 & Badger & $144.43$ &  9 \\ 
7 & Kerrtown & $124.74$ &  7 \\ 
8 & Chireno & $152.69$ &  9 \\
9 & Calavo & $239.24$ & 15 \\
10 & Auburn & $430.31$ & 25 \\
11 & Lawson & $1859.6$ &  - \\
12 & HAAS & $843.18$ &  - \\
\bottomrule
\end{tabular}}
\caption{List of our simulated and real-world environments along with their dimensions and number of rooms. The last two are real-world environments containing hallways and student common areas with no concrete partitions.} 
\label{table:env_info}
\vspace{-0.01in}\end{table}

\begin{table}[h]\vspace{0.1in}
\centering
\scalebox{1}{ 
\begin{tabular}{cccc}
\toprule
\multirow{2}{*}{Methods} & \multicolumn{3}{c}{Performance Metrics} \\ 
\cmidrule{2-4}
& \multicolumn{1}{c}{Time (sec) $\downarrow$} & \multicolumn{1}{c}{Length $\downarrow$} & \multicolumn{1}{c}{SR (\%) $\uparrow$} \\ 
\midrule
Ours  & $0.057 \pm 0.018$ & $4.52 \pm 0.51$ & 97.4 \\
NTFields  & $0.014 \pm 0.002$ & $3.37 \pm 0.54$ & 65.0 \\
P-NTFields & $0.012 \pm 0.001$ & $3.46 \pm 0.57$ & 65.8 \\
RRTConnect & $1.918 \pm 1.738$ & $4.15 \pm 0.76$ & 100.0 \\
LazyPRM & $0.381\pm 0.261$ & $4.35 \pm 0.65$ & 98.8 \\
FMM & $0.822 \pm 0.028$ & $4.70 \pm 0.46$ &  100.0 \\ 
\bottomrule
\end{tabular}}
\caption{Comparison for all motion planning methods in ten Gibson environments. It can be seen that our method exhibits higher SR than other PiNMPs but relatively slower planning times. The relatively slow planning times are because our method solves much harder cases with start and goal spanning across multiple rooms. }
\label{table:mp_gib}
\vspace{-0.15in}\end{table}

\begin{figure*}[t]
\centering
\begin{subfigure}{0.48\textwidth}
\centering
\includegraphics[trim={0cm 4cm 6cm 5cm}, clip,width=1.0\linewidth]{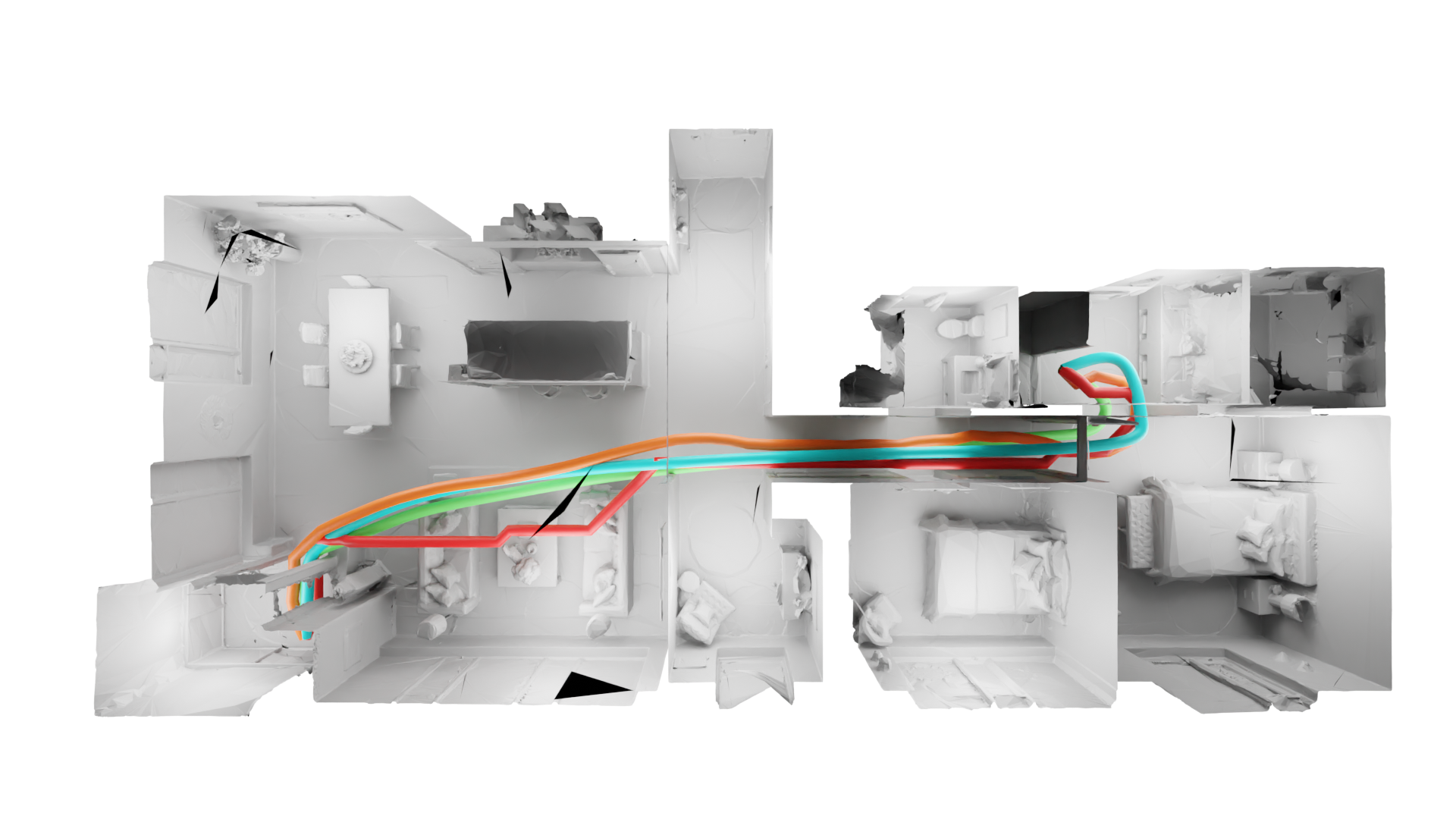}
\caption{{Chireno: 152.69 Sq. m}}
\end{subfigure}
\begin{subfigure}{0.48\textwidth}
\centering
\includegraphics[trim={0cm 4cm 6cm 1cm}, clip,width=1.0\linewidth]{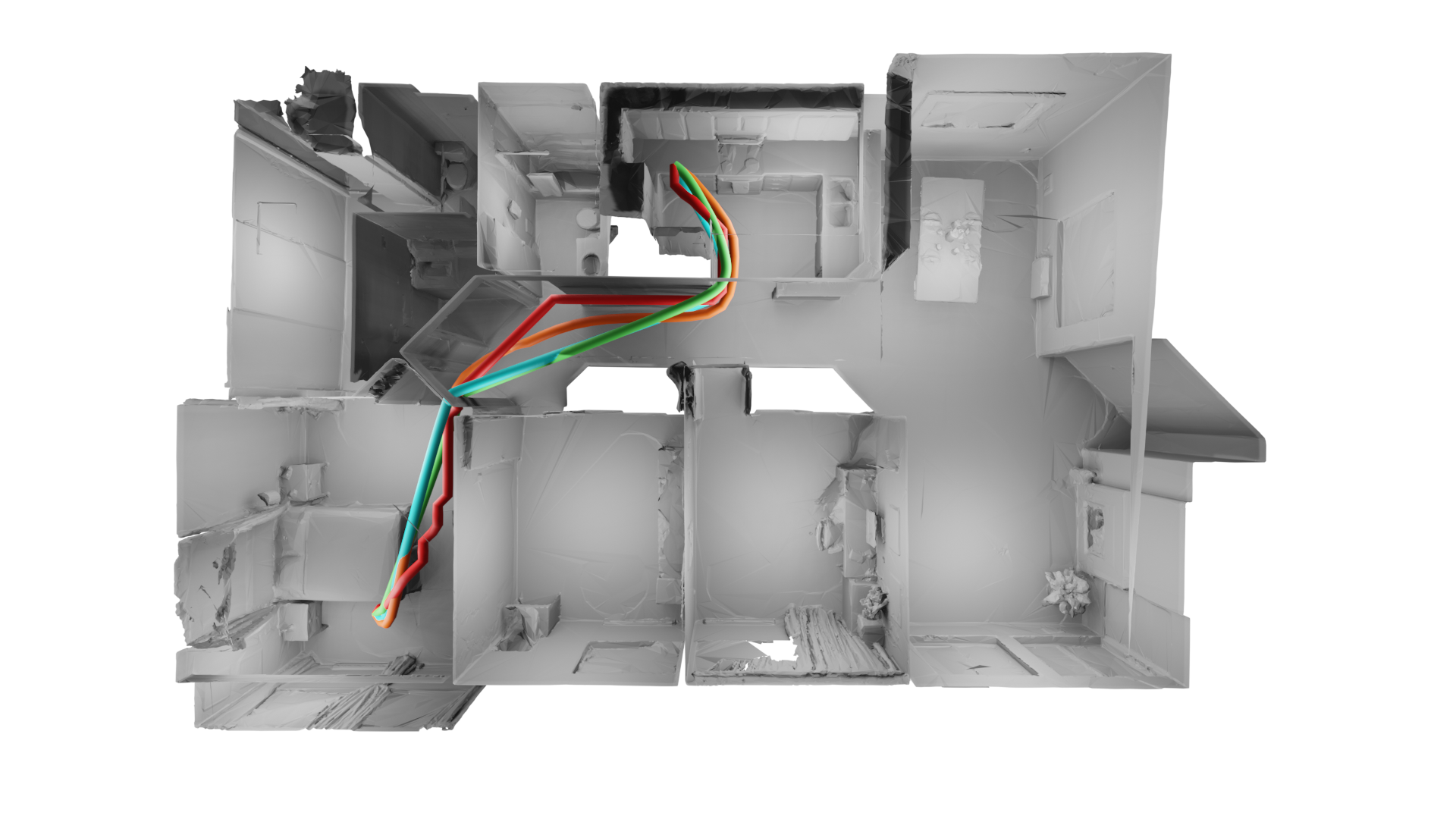}
\caption{Hercules: 119.16 Sq. m}
\end{subfigure}
\caption{Depiction of two Gibson environments: The figure illustrates the paths generated by each method between a same start and goal. NTFields and P-NTFields failed to produce viable paths, while our method successfully inferred smooth paths (orange) in approximately 0.12 seconds, showcasing its ability to operate in high-frequency, large domains. In comparison, the SMP methods RRTConnect (green) and Lazy-PRM (cyan) required an average of 3.80 and 2.12 seconds, respectively, and produced paths with sharp turns. FMM (red) generated paths in 0.94 and 0.97 seconds; however, these paths exhibit numerous turns due to the grid search.}
\label{gibson}\vspace{-0.1in}
\end{figure*}

\subsection{Comparison Analysis over Simulated Environments}
\textbf{Baselines.} We consider the following five baseline approaches, including NTFields \cite{ni2023ntfields}, P-NTFields \cite{ni2023progressive}, RRTConnect with smoothing \cite{kuffner2000rrt}, LazyPRM with smoothing \cite{bohlin2000path}, and the FMM \cite{sethian1996fast}. NTFields and P-NTFields represent state-of-the-art PiNMPs that do not require expert training data. RRTConnect and LazyPRM are known classical sampling-based methods. Finally, the FMM method also solves Eikonal PDE like ours but relies on discretization and search to infer the arrival time fields. All motion planning algorithms use the KDTree \cite{grandits_geasi_2021} as a collision check to ensure fairness. 

\textbf{Metrics.} The evaluation metrics include motion planning time, path length, and success rate (SR). A time limit of 10 seconds is imposed for classical methods. If a planner cannot retrieve a path within that limit, that case will be considered a failure. Planning time and path length are only recorded for successful cases. 

\textbf{Simulation Experiments.} In these experiments, we compare all methods in the Gibson environments which present cluttered home-like indoor scenarios. Table. \ref{table:env_info} summarize our ten environments' dimensions and their total number of rooms.  We sampled 200 start and goal positions for each environment to evaluate motion planning performance.

In these environments, our method, NTFields, and P-NTfields took, on average, {9}, {18}, and {71} minutes for training, respectively. The LazyPRM took {1} minutes on average for graph construction. Table~\ref{table:mp_gib} summarizes the results of all methods and demonstrates the success rate of all methods across each environment. Note that the planning times do not include the graph construction times for the LazyPRM.

Table~\ref{table:mp_gib} shows that existing PiNMPs, such as NTFields and P-NTFields, offer rapid inference times but suffer from much lower success rates compared to our methods. In contrast, traditional sampling-based methods like RRTConnect and LazyPRM, as well as the grid-based FMM, maintain high success rates yet with longer planning times. In terms of path length, NTFields and P-NTFields can only manage shorter paths, while sampling-based methods often produce shorter paths that are near obstacles. FMM, similar to our approach, computes paths based on travel times, but its grid-search strategy tends to result in longer paths. As Fig.~\ref{gibson} illustrates, sampling-based methods generate trajectories with abrupt turns, whereas FMM’s paths, affected by grid discretization, display numerous small turns.

Overall, the results confirm that our method scales efficiently to complex Gibson environments. It completes motion planning in less than 0.1 seconds while maintaining a high success rate across all environments.
\begin{table}[h]
\centering
\scalebox{1}{ 
\begin{tabular}{cccc}
\toprule
\multirow{2}{*}{Methods} & \multicolumn{3}{c}{Performance Metrics} \\ 
\cmidrule{2-4}
& \multicolumn{1}{c}{Time (sec) $\downarrow$} & \multicolumn{1}{c}{Length $\downarrow$} & \multicolumn{1}{c}{SR (\%) $\uparrow$} \\ 
\midrule
Ours  & $0.092 \pm 0.018$ & $4.36 \pm 1.22$ & 97.0 \\
RRTConnect & $2.017\pm 1.179$ & $4.55 \pm 1.41$ & 100.0 \\
LazyPRM & $0.715\pm 0.061$ & $4.13 \pm 1.40$ & 100.0 \\
\bottomrule
\end{tabular}}
\caption{Comparison of our method and classical approaches in real-world settings. NTFields and P-NTFields failed to converge in those settings and are excluded from the comparison. Note that the classical method planning time increased significantly as the domain size became larger, whereas our method retains its high performance.} 
\label{table:real}
\vspace{-0.05in}\end{table}

\subsection{Comparison Analysis over Real-world Environments}

Our real-world experiments demonstrate that our method can effectively scale to complex, real-world scenarios. These experiments were conducted in two distinct settings: HAAS, which contains long, narrow passage halls, and Lawson, which contains passages and cluttered study areas. The HAAS environment has dimensions of $843.18$ Sq. meters and is shown in Fig.~\ref{fig:haas}. While Lawson covers $1859.6$ Sq. Meters, and its map is depicted in Fig.~\ref{fig:pipeline}. In these domains, we sample 200 start and goal pairs in each environment for evaluation. The experiments utilized a Unitree B1 quadruped robot equipped with a PandarXT-16 LiDAR sensor for perception. The demonstration videos of a robot dog navigating HAAS and Lawson environments using our method are provided with supplementary material. 

In these real-world scenarios, we pick the sampling-based methods with the consideration of avoiding multiple small turns. The NTFields and P-NTFields are also excluded because they failed to converge. The FB-NTFields training times in HAAS and Lawson were {36} and {6} minutes, respectively. The Lazy PRM took {4} minutes to construct the graph. Table. \ref{table:real} compares all metrics of our method with classical MP. It can be seen that our method retains its high performance with high SR and low planning times. The classical methods' planning time increased significantly due to the larger domain size. These results demonstrate that our method is efficient for path inference, maintaining a high success rate, and handling real-world environments with varying complexity. Finally, Fig.~\ref{fig:haas} depicts a robot dog navigating a real-world HAAS environment using the planned path by our approach. Additionally, Fig.~\ref{fig:pipeline} shows the arrival time fields by our approach in Lawson. 

\section {Conclusions and Future Work}


Neural Time Fields provide a data-efficient approach to solving the Eikonal PDE for motion planning without requiring demonstration data. However, existing Physics-informed Neural Motion Planning methods struggle with scalability, as high-frequency features become more prevalent in larger environments, exacerbating spectral bias and optimization challenges. To address these limitations, we propose FB-NTFields, which decompose the domain into overlapping subdomains and learn a latent space distance representation, enabling efficient arrival time field inference for complex motion planning tasks. Our results demonstrate that FB-NTFields outperform prior methods across simulated and real-world large-scale environments.

Looking ahead, we aim to extend FB-NTFields to mobile manipulation tasks in high-dimensional configuration spaces. Future work will explore decomposing along both the robot's base and manipulation dimensions, allowing the composition of features for solving the Eikonal PDE in joint space. Additionally, we seek to scale our method to dynamic, large indoor and outdoor environments where high-frequency variations arise more frequently. This requires adaptive decomposition strategies capable of dynamically updating arrival time fields to navigate long-horizon goals while avoiding moving obstacles.

\bibliographystyle{IEEEtran}
\bibliography{references}

\end{document}